\documentclass[letterpaper, 10 pt, conference]{ieeeconf}
\IEEEoverridecommandlockouts 
\usepackage{cite}
\usepackage{amsmath,amssymb,amsfonts}
\usepackage{algorithmic}
\usepackage{graphicx}
\usepackage{textcomp}
\usepackage{multirow}
\usepackage{arydshln}
\usepackage{xcolor}
\def\BibTeX{{\rm B\kern-.05em{\sc i\kern-.025em b}\kern-.08em
    T\kern-.1667em\lower.7ex\hbox{E}\kern-.125emX}}

\title{\LARGE \bf
Pedestrian Intention Prediction via Vision-Language Foundation Models\\
}

\author{Mohsen Azarmi$^{1}$, Mahdi Rezaei$^{1,\dag}$, and He Wang$^{2}$
\thanks{$^{1}$M. Azarmi and M. Rezaei are with the Institute for Transport Studies, Faculty of Environment, Computer Vision and Machine Learning Group, University~of~Leeds, LS2 9JT, United Kingdom
        {\tt tsmaz@leeds.ac.uk} and {\tt  m.rezaei@leeds.ac.uk} }%
\thanks{$^{2}$H. Wang is with AI Centre, Department of Computer Science, University College London, London, WC1E 6BT, United Kingdom
       {\tt he\_wang@ucl.ac.uk}}%
\thanks{$^{\dag}$Corresponding author: M. Rezaei {\tt m.rezaei@leeds.ac.uk}}
}

\begin{document}

\maketitle
\thispagestyle{empty}
\pagestyle{empty}

\begin{abstract}
Prediction of pedestrian crossing intention is a critical function in autonomous vehicles. Conventional vision-based methods of crossing intention prediction often struggle with generalizability, context understanding, and causal reasoning. This study explores the potential of vision-language foundation models (VLFMs) for predicting pedestrian crossing intentions by integrating multimodal data through hierarchical prompt templates. The methodology incorporates contextual information, including visual frames, physical cues observations, and ego-vehicle dynamics, into systematically refined prompts to guide VLFMs effectively in intention prediction. Experiments were conducted on three common datasets—JAAD, PIE, and FU-PIP. Results demonstrate that incorporating vehicle speed, its variations over time, and time-conscious prompts significantly enhances the prediction accuracy up to 19.8\%. Additionally, optimised prompts generated via an automatic prompt engineering framework yielded 12.5\% further accuracy gains. These findings highlight the superior performance of VLFMs compared to conventional vision-based models, offering enhanced generalisation and contextual understanding for autonomous driving applications.
\end{abstract}


\section{Introduction}
The safe and efficient operation of autonomous vehicles (AVs) relies on understanding pedestrian crossing intentions. 
Accurate intention prediction enhances safety by enabling AVs to anticipate actions and adjust their speed and trajectory accordingly. However, this task remains challenging due to the complexity of pedestrian behaviour, influenced by individual attributes, social interactions, and environmental factors \cite{rasouli2019autonomous}.

Conventional computer vision-based models for crossing intention prediction often use deep learning techniques, such as CNNs \cite{razali2021pedestrian,zhang2022st,azarmi2024pip}, RNNs \cite{rasouli2020PedestrianAA,kotseruba2020they,rasouli2022multi}, GCNs \cite{cadena2022pedestrian,chen2021visual}, and Transformers \cite{Lorenzo2021CAPformerPC,zhou2023pit,sharma2025predicting}, focusing on visual features like body pose \cite{yang2021pedestrian,ahmed2023multi}, spatio-temporal relationships \cite{liu2020spatiotemporal,saleh2019real}, and pedestrian-vehicle dynamics \cite{xu2024pedestrian}. While effective, these models struggle with generalizability, context understanding, and causal reasoning in dynamic traffic environments \cite{zhang2024causal}. 

\begin{figure}
    \centerline{\includegraphics[width=2.6in]{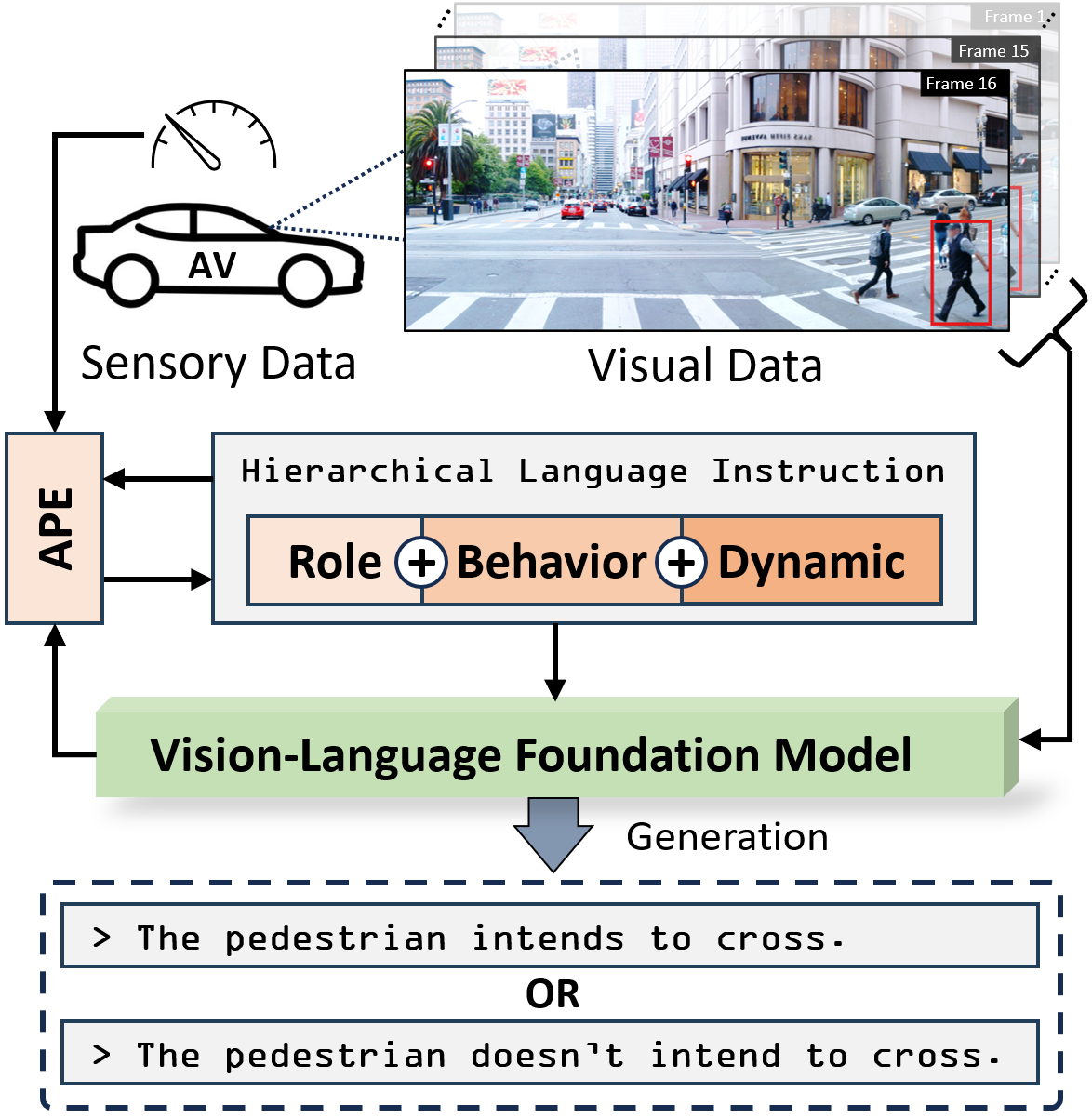}}
    \vspace{-3mm}
    \caption{Overview of vision-language crossing intention prediction model. Automatic Prompt Engineer (APE) generates hierarchical prompts based on task role, pedestrian behaviour, and dynamic motion cues (e.g., vehicle speed). \label{fig:vlm}}
    \vspace{-4mm}
\end{figure}

Vision-language foundation models (VLFMs) are large-scale, pre-trained machine learning models that integrate visual and textual modalities to achieve a unified understanding of multimodal data. By combining multimodal data, VLFMs address some limitations of conventional vision-based models, improving cross-modal comprehension tasks \cite{zhou2024vision}. 
While some recent works have explored the use of VLFMs for pedestrian intention prediction tasks \cite{huang2024gpt,hamomnipredict,munirpedvlm}, they exhibit key limitations. Firstly, these approaches often fail to effectively capture temporal dependencies critical for dynamic pedestrian behaviour, relying instead on static or limited contextual cues. Secondly, their performance is sensitive to prompt variations, as prompts are often designed in an arbitrary or ad hoc manner, lacking a systematic structure. Finally, these studies frequently overlook the significant role of vehicle dynamics—such as speed and acceleration—on pedestrian intent, which are crucial for understanding traffic scenes 
\cite{Lorenzo2021CAPformerPC,zhang2023cross,azarmi2024feature}.

This study explores the use of four state-of-the-art VLFMs for pedestrian crossing intention prediction, with a particular focus on prompt engineering and the integration of vehicle speed variations and time-conscious prompts into the intent-predictive framework. 

This research introduces structured hierarchical templates for prompt development. The hierarchy progresses from simpler to more complex prompts, each designed to capture different levels of contextual information. This progressive approach reflects human reasoning, which often builds on foundational knowledge before addressing more intricate details \cite{budagam2024hierarchical}. These prompts are subsequently optimised using the Automatic Prompt Engineer (APE) framework \cite{zhou2022large}, which iteratively refines them to identify the high-score prompts. The study also investigates how varying levels of contextual information influence crossing intention prediction performance against three task-specific datasets.
Then our contributions can be summarised as follows:
\begin{itemize}
    \item \textbf{VLFMs Adaptation:} We integrate vehicle speed variations and refine prompts to improve the understanding of temporal and physical cues in pedestrian intention prediction. 

    \item \textbf{Hierarchical Prompt Design:} We propose a structured series of progressively complex prompts to evaluate VLFMs' predictive capabilities.

    \item \textbf{Prompt Optimisation:} We employ the APE framework to systematically optimise prompt templates, balancing accuracy and confidence, and improving the overall performance of the VLFMs in pedestrian crossing intention prediction.

\end{itemize}

\section{Methodology}
This section describes our approach to pedestrian crossing intention prediction using vision-language foundation models (VLFMs). We integrate vehicle dynamics into the prompt structure through hierarchical templates and optimise prompts to enhance VLFMs' effectiveness for the task.

\subsection{Input Data Preparation}

The data we utilised to input to the VLFMs comprises both visual and textual components as detailed below.

\subsubsection{Visual Data}
The visual data consists of a series of frames, $\mathbf{V}_i = {frame_1, frame_2, \ldots, frame_N}$, with $N=16$ frames captured at a rate of 30 fps. Following the same framework employed by other conventional vision-based benchmark models for performance evaluation, the model is configured to analyse information from the previous 16 frames and generate predictions for the next 16 frames \cite{kotseruba2021benchmark}. To guide the model in understanding temporal sequences, each frame includes annotations with timestamps and a red bounding box highlighting the target ($i$-th) pedestrian (as illustrated in Figure \ref{fig:vlm}).

\subsubsection{Textual Data}
The textual data consists of a set of prompt templates $\mathbf{P} = \{\rho_1, \rho_2, \ldots, \rho_n\}$. These prompts direct the VLFMs to predict the pedestrian’s crossing intention based on visual frames and prompts, which are detailed below.

\subsection{Hierarchical Prompt Templates}

To systematically evaluate the impact of contextual richness on the performance of VLFMs in pedestrian crossing intention prediction, we propose a series of progressively complex prompt templates as suggested by \cite{budagam2024hierarchical}. Each level of these templates builds upon the previous one, incrementally incorporating additional information to guide the model more effectively. This hierarchical structure ensures a comprehensive analysis of how varying degrees of context affect the model's understanding of the scene. 

\subsubsection{Role Templates ($\mathbf{P}_{R}$)} The system role initially is based on definitions suggested in the study \cite{hamomnipredict}, where the model acts as an autonomous vehicle equipped with a front-view dashboard camera capturing traffic scenes, including pedestrians and vehicles. The task requires predicting a pedestrian's behaviour 16 frames into the future based on visual inputs and ego-vehicle speed information. The defined role helps the VLFM focus on the task effectively, as supported by \cite{kong2023better}. 

The baseline prompt poses a simple question to establish a control for performance evaluation: \textit{"Does the pedestrian in the red bounding box intend to cross the street?"}. This minimal template serves as a benchmark for comparison with more contextually rich prompts. 
Moreover, referring to the bounding box in the prompt reduces ambiguity by ensuring the model targets the correct individual. Providing this localised visual reference enables the model to focus on the pedestrian more accurately, avoiding inaccuracies that may arise from textual descriptions of bounding box coordinates \cite{zhang2024good}.

\subsubsection{Physical Cues Observation ($\mathbf{P}_{B}$)} This adds a layer of behavioural insight by specifying \textit{"Observe the pedestrian's posture, limb positions, and body orientation"}. By explicitly focusing on these micro-level features, this template provides critical insights into the pedestrian's readiness and intent to cross \cite{fang2018pedestrian}.

\subsubsection{Vehicle Dynamics Aware} The template integrates ego-vehicle dynamics, incorporating vehicle speed as a key contextual element in pedestrian intention prediction \cite{azarmi2024feature}.
Three different representations of vehicle speeds are designed to evaluate the effect of varying levels of detail in dynamic information on intention prediction. The first representation ($\mathbf{P}_{Ds}$) provides a straightforward numeric value, offering the model a precise but contextually limited input. The second representation ($\mathbf{P}_{Dd}$) adds interpretability by describing the vehicle's motion state qualitatively, helping the model infer broader situational awareness, as used by \cite{hamomnipredict}. Finally, the third representation ($\mathbf{P}_{Dt}$) introduces a time-conscious element, enabling the model to consider dynamic changes in vehicle speed over a given time interval, which closely aligns with real-world traffic scenarios. For example: \textit{"Over the past \{time interval\} seconds, the vehicle's speed \{increased / decreased\} from \{initial speed\} mph to \{final speed\} mph."}

\subsection{Prompt Optimization}

To optimise prompt templates, we use the Automatic Prompt Engineer (APE) framework \cite{zhou2022large}, which systematically generates and evaluates prompts. Each prompt $\rho_j$ is scored using:
\[
f_{\text{score}}(\rho_j) = \alpha f_{\text{exec}}(\rho_j) + (1 - \alpha) f_{\text{logprob}}(\rho_j),
\]
where:
\begin{itemize}
    \item Execution Accuracy:
    \vspace{-2mm}
    \[
    f_{\text{exec}}(\rho_j) = \frac{1}{M} \sum_{i=1}^M \mathbb{I}(\hat{y}_i = y_i),
    \]
    measures the proportion of correct predictions, where $\mathbb{I}(\cdot)$ is the indicator function, $y_i$ is the ground-truth label, $\hat{y}_i$ is the predicted label, and $M$ is the number of samples.
    \item Log Probability:
    \vspace{-2mm}
    \[
    f_{\text{logprob}}(\rho_j) = \frac{1}{M} \sum_{i=1}^M \log P_{\text{VLFM}}(y_i \mid \rho_j, \mathbf{V}_i),
    \]
    computes the average confidence of the VLFM for the correct label $y_i$, given the sequence of images $\mathbf{V}_i$ and the prompt $\rho_j$.
\end{itemize}
The weighting parameter $\alpha \in [0, 1]$ balances accuracy and confidence.

The employed Monte Carlo search in the framework iteratively refines prompts:
\begin{enumerate}
    \item Initialisation: Start with the initial set of templates ($\mathbf{P}$).
    \item Perturbation: Generate new candidate prompts (via ChatGPT \cite{OpenAI2023ChatGPT}) by modifying existing ones, such as rephrasing, or using synonyms. 
    \item Evaluation: Score the new prompts using $f_{\text{score}}(\rho_j)$ to identify high-performing candidates.
    \item Selection: Retain the top $K$ prompts for the next iteration.
\end{enumerate}
This process repeats for $T$ iterations or until convergence, yielding the optimized prompt $\rho^*$:
\[
\rho^* = \arg\max_{\rho_j} f_{\text{score}}(\rho_j).
\]

The optimised prompt $\rho^*$, which is both linguistically diverse and contextually relevant, is used as the input query for the VLFM to predict pedestrian crossing intentions effectively.

\section{Experiment}

This section details the experimental results of assessing the impact of hierarchical prompt templates and the incorporation of vehicle speed variations on the performance of VLFMs using the APE framework. Then, it provides comparative results on three task-specific datasets using four state-of-the-art VLFMs and conventional vision-based intent-predictive models.

\subsection{Datasets}
Our study utilizes three benchmark datasets: JAAD \cite{Rasouli2017IV}, PIE \cite{rasouli2019pie}, and Frontal Urban-PIP (FU-PIP) \cite{azarmi2024pip}, each providing diverse pedestrian scenarios for intention prediction. For all datasets, both test and validation sets are fully annotated with pedestrian intention labels.  

JAAD consists of 126 test samples and 32 validation samples, including behavioural annotations essential for pedestrian intention prediction. The dataset captures varied weather conditions (sunny, rainy, snowy) and unpredictable pedestrian movements, such as sudden stops or changes in direction, increasing task complexity.  

PIE includes 719 test samples and 243 validation samples, covering diverse urban environments like motorways, parking lots, and intersections. Frequent occlusions and dynamic interactions with vehicles introduce additional challenges for intention prediction models.  

Frontal Urban-PIP (FU-PIP) contains 94 test samples and 90 validation samples, featuring complex traffic interactions at intersections with high vehicle flow. The dataset evaluates the model's ability to recognise subtle pedestrian intent cues in busy urban settings.  

All datasets consist of RGB video sequences recorded from frontal vehicle-mounted cameras, annotated with pedestrian bounding boxes, ego-vehicle speeds, and binary intention labels ("Crossing" or "Not Crossing"). 

\subsection{Hierarchical Optimization}

To explore the effectiveness of the hierarchical prompt templates, we utilize five distinct language instruction pools, which include two pools for Role Templates ($\mathbf{P}_{R}$) and Physical Cues Observation Templates ($\mathbf{P}_{B}$), and three for Vehicle Dynamics ($\mathbf{P}_{Ds}$, $\mathbf{P}_{Dd}$, $\mathbf{P}_{Dt}$). Each of these pools contains placeholders for speed information to guide the VLFMs in predicting pedestrians' crossing intentions.

Each instruction pool initially includes 13 manually crafted templates, inspired by \cite{huang2024gpt,hamomnipredict,munirpedvlm} prompts. These templates are modified and developed to ensure diversity in grammar, structure, and lexical richness while maintaining the original semantic meaning. This diversity ensures the prompts effectively probe the capabilities of the models under varied linguistic constructions.

We employ the APE framework using GPT-4V for task evaluation and leverage ChatGPT-4 \cite{achiam2024gpt} to refine and optimise these templates. The optimisation process is conducted with $M=365$ (number of samples) and $T=40$ (number of iterations) across the validation sets of all three datasets.
Optimising the prompt templates across these three datasets ensures that the prompts are effective and well-tuned for diverse scenarios, covering various environments, pedestrian behaviours, and levels of complexity.

The optimisation procedure involves the following stages:
\begin{enumerate}
    \item Role Templates ($\mathbf{P}_{R}$):
     Using the validation samples, we first run APE on the Role Templates to identify the most effective template for guiding the VLFMs. From this step, we select the top 5 ($K$=5) high-score prompts for further analysis.

    \item Physical Cues Observation Templates ($\mathbf{P}_{B}$):
    Building upon the top 5 prompts from $\mathbf{P}_{R}$, we evaluate the observation  templates ($\mathbf{P}_{B}$). This step incorporates behavioural cues into the prompts to enhance the contextual understanding of pedestrian actions. The optimisation process yields the most effective prompt for this pool.

    \item Vehicle Dynamics Templates ($\mathbf{P}_{Ds}$, $\mathbf{P}_{Dd}$, $\mathbf{P}_{Dt}$):
    Finally, using the top 5 high-score prompts from $\mathbf{P}_{B}$, we evaluate the three representations of ego-vehicle speed to determine their contribution to the overall performance. However, due to the limitations of the JAAD dataset, which only provides descriptive speed information, the experiments on this dataset are restricted to descriptive speed templates ($\mathbf{P}_{Dd}$).
\end{enumerate}

\begin{figure}
    \centerline{\includegraphics[width=3.3in]{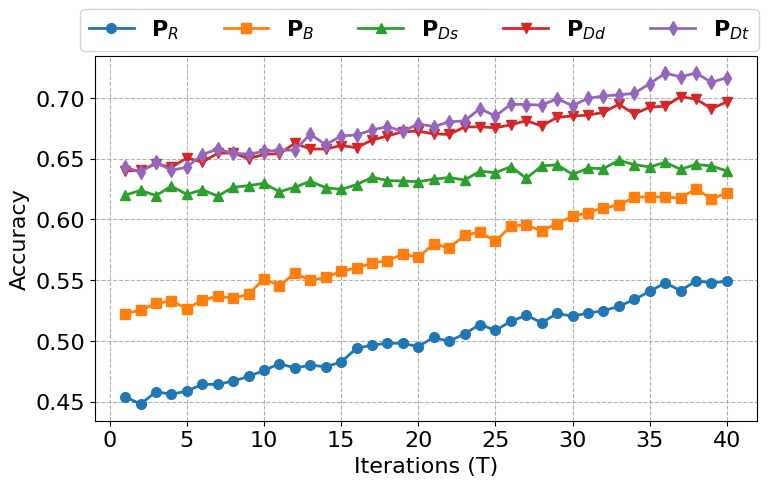}}
    \vspace{-2mm}
    \caption{The measured performance of GPT-4V during the prompt optimisation process of hierarchical prompt templates on validation sets of JAAD, PIE, and FU-PIP datasets. 
    \label{fig:optim}}
    \vspace{-3mm}
\end{figure}

Figure \ref{fig:optim} shows the optimisation process of prompt templates, with stability achieved after iteration 34, indicating convergence on the validation sets and the identification of the best-performing templates for pedestrian crossing intention prediction across three task-specific datasets. 
Notably, including the ego-vehicle's speed value alone resulted in a +2\% improvement in validation performance. In contrast, both descriptive speed and time-conscious descriptive speed showed a similar trend prior to iteration 20, with subsequent refinements yielding accuracy improvements of +8\% and +10\%, respectively. This emphasises the importance of temporal and dynamic context in scene understanding. 

Table \ref{tab:optim} represents the evaluation of the best-performing prompt in each template pool using accuracy, F1 score, and log probability. As the complexity of the templates increases, both accuracy and F1 scores improve, while log probabilities decrease, indicating more confident predictions. The $\mathbf{P}_{Dt}$ template, which includes time-conscious vehicle dynamics, performs best with an accuracy of 0.72, F1 score of 0.71, and a log probability of -1.52. In contrast, the $\mathbf{P}_R$ template, with minimal context, performs the worst with an accuracy of 0.54, an F1 score of 0.51, and a log probability of -2.05.
\vspace{-3mm}
\begin{table}[ht]
    \centering
    \caption{Performance Evaluation for Different Prompt Templates.}
    \vspace{-2mm}
    \begin{tabular}{c|c|c|c}
    \noalign{\hrule height 1.2pt}
    \textbf{Prompt Template} & \textbf{Accuracy $\uparrow$} & \textbf{F1 Score $\uparrow$} & \textbf{Log Probability $\downarrow$} \\
    \hline
    \textbf{$\mathbf{P}_R$} & 0.54 & 0.51 & -2.05 \\
    \textbf{$\mathbf{P}_B$} & 0.62 & 0.60 & -1.92 \\
    \textbf{$\mathbf{P}_{Ds}$} & 0.64 & 0.62 & -1.60 \\
    \textbf{$\mathbf{P}_{Dd}$ } & 0.70 & 0.69 & -1.54 \\
    \textbf{$\mathbf{P}_{Dt}$ } & 0.72 & 0.71 & -1.52 \\
    \noalign{\hrule height 1.2pt}
    \end{tabular}
    \label{tab:optim}
\end{table}
\vspace{-3mm}
To address potential bias toward GPT-4V, we extended the optimisation process to other VLFMs, including GPT-4 mini\cite{zhu2023minigpt}, LLaVA-Next (7B) \cite{li2024llava}, and LLaVA-Next (3B) \cite{li2024llava}. Specifically, we continued refining the optimised prompt templates on these models for an additional 10 iterations. However, this extended tuning process did not yield any improvements on the top-10 high-scoring prompts, indicating that the optimised templates were already well-aligned with the models' capabilities. This suggests that the observed performance differences across models stem from inherent variations in their architectures rather than any prompt-specific bias favouring GPT-4V.

\subsection{Effectiveness of Prompts and Structures}

We evaluate the utility and relevance of high-performing and low-performing keywords, phrases, and grammar structures based on experimental results.

\subsubsection{High-Performing Keywords and Phrases}
Our analysis reveals that prompts incorporating specific behavioural, contextual, and temporal keywords consistently achieve higher prompt scores. For example, keywords such as \textit{“posture”}, \textit{“movement”}, and \textit{“orientation”} were instrumental in directing the model’s attention to observable pedestrian features. Similarly, terms like \textit{“road”}, \textit{“crosswalk”}, \textit{“traffic light”}, and \textit{“proximity to the”} effectively framed the spatial relationships between the pedestrian and their environment, providing crucial context for intention prediction.

Temporal references, such as \textit{“over the last \{X\} seconds”} and \textit{“in the \{X\} frames”}, enabled the model to analyse dynamic changes in behaviour, enhancing its ability to predict intentions accurately. Additionally, phrases that explicitly described vehicle dynamics, including \textit{“ego-vehicle speed”}, \textit{“slowing down”}, and \textit{“deceleration”}, facilitated a better understanding of interactions between the vehicle and the pedestrian. For instance, prompts like \textit{“given the ego-vehicle’s deceleration, consider the pedestrian’s movement and proximity to the crosswalk”} consistently yielded high performance, specifically on the FU-PIP dataset, which over presents the designated intersection scenarios.

From a grammatical perspective, direct and concise question structures proved to be ineffective. On the other hand, the integration of sub-problems seems to provide chain-of-thought reasoning, standing as the most effective template. For instance, a prompt such as \textit{``First, observe the pedestrian in the red bounding box and their proximity to the road over the sequence of 16 given frames. Then, consider pedestrian’s movements and posture across frames, and determine if there is enough evidence for the pedestrian to cross the road"}. This stepwise breakdown encourages the model to focus on key factors sequentially. This aligns with \cite{budagam2024hierarchical}, where breaking the task into sub-problems helps the VLFM focus on granular aspects of behaviour before forming a conclusion.
In the second place, comparative structures such as \textit{“Compare the pedestrian’s movements in the first and last frames”} encouraged the model to analyse temporal dependencies, further improving prediction accuracy in JAAD and PIE datasets.

\subsubsection{Low-Performing Keywords and Phrases}
Prompts containing vague or ambiguous keywords demonstrated significantly lower prompt scores. Terms like \textit{``acting"} and \textit{``behaving"}, lacked the specificity needed to guide the model effectively. Similarly, abstract descriptors such as \textit{``tendency"}, \textit{``desire"}, and \textit{"feel"} introduce uncertainty, as they imply a degree of probability rather than a definitive observation, which hinders the model's ability to generate accurate predictions.

Redundant or irrelevant keywords, such as \textit{“background elements”} and \textit{“objects around”}, diluted the model’s focus on pedestrian behaviour, negatively impacting the prompt score. Furthermore, non-actionable temporal references like \textit{“in the past”} failed to provide sufficient temporal context, reducing their utility in dynamic scenes.

In terms of grammar structures, overly complex sentences were found to hinder the model’s comprehension. For instance, a prompt like \textit{“Analyse the pedestrian’s movements, body pose, and posture to determine whether they are likely to cross, but also take into account the vehicle is moving fast}” overloaded the model with information and reduced prediction accuracy. Similarly, the use of passive voice, as in \textit{“The intention to cross is being assessed”}, was less effective than active constructions. Open-ended or unstructured prompts, such as \textit{“What do you think about the pedestrian’s behaviour?”}, also performed poorly due to their lack of focus on the task.

\subsection{Comparison Results}

Using the optimized prompts, we evaluate four VLFMs, including GPT-4V (1.8T) \cite{achiam2024gpt}, GPT-4 mini (8B) \cite{zhu2023minigpt}, LLaVA-Next (7B) \cite{li2024llava}, and LLaVA-Next (3B) \cite{li2024llava}, against the test sets of the three datasets. 
Each dataset presents a variety of real-world scenarios, ranging from urban streets to residential areas, with varying levels of complexity in terms of pedestrian behaviour and environmental factors.

The average results of the top 5 highest-scoring optimized prompts are presented in Table \ref{tab:vlfm_comparison}, using task-specific metrics proposed by \cite{kotseruba2021benchmark}, such as Accuracy (Acc), Area Under the Curve (AUC), F1-score (F1), Precision (Pr), and Recall (Re), for each model and dataset.

Vision-based models, such as MultiRNN, SingleRNN, and StakedRNN, show strong performance on the PIE dataset, with MultiRNN achieving the highest precision and recall on JAAD (0.64 and 0.86, respectively). However, these models show relatively poor results on the FU-PIP dataset, indicating their limited ability to handle more complex or diverse pedestrian behaviours. PCPA and GraphPlus, while performing well on PIE (0.87 Acc, 0.86 AUC, 0.77 F1 for PCPA), struggle on FU-PIP, where their accuracy drops to around 0.62 and F1 remains low. PIP-Net outperforms other vision-based models, achieving the highest accuracy and F1 score on PIE (0.91 Acc, 0.84 F1) and FU-PIP (0.73 Acc, 0.69 F1), highlighting its robustness across datasets.

\begin{table}[ht]
    \centering
   \vspace{3mm}
    \caption{Comparison of VLFMs on Test Sets of JAAD, PIE, and FU-PIP.}
    \vspace{-3mm}
    \setlength{\tabcolsep}{4pt} 
    \begin{tabular}{ccccccccc}
    \noalign{\hrule height 1.2pt}
    \textbf{Model} & \textbf{Dataset} & \textbf{Acc} & \textbf{AUC} & \textbf{F1} & \textbf{Pr} & \textbf{Re} \\
    \noalign{\hrule height 1.2pt}
    \multicolumn{7}{c}{\textbf{Vision-Based Models}} \\
    \hline
    \multirow{3}{*}{MultiRNN \cite{rasouli2022multi}} 
    & JAAD & 0.61 & 0.50 & \textcolor{blue}{0.74} & 0.64 & \textcolor{blue}{\textbf{0.86}} \\
    & PIE & 0.83 & 0.80 & 0.71 & 0.69 & 0.73 \\
    & FU-PIP & 0.64 & 0.63 & 0.49 & 0.51 & 0.48 \\
    \hdashline
    \multirow{3}{*}{SingleRNN \cite{kotseruba2020they}} 
    & JAAD & \textcolor{blue}{0.58} & \textcolor{blue}{0.54} & 0.67 & \textcolor{blue}{0.67} & 0.68 \\
    & PIE & 0.81 & 0.75 & 0.64 & 0.67 & 0.61 \\
    & FU-PIP & 0.65 & 0.64 & 0.54 & 0.57 & 0.53 \\
    \hdashline
    \multirow{3}{*}{StakedRNN \cite{rasouli2020PedestrianAA}} 
    & JAAD & \textcolor{blue}{0.58} & \textcolor{blue}{0.54} & 0.69 & \textcolor{blue}{0.67} & 0.61 \\
    & PIE & 0.82 & 0.79 & 0.69 & 0.67 & 0.70 \\
    & FU-PIP & 0.65 & 0.65 & 0.55 & 0.58 & 0.53 \\
    \hdashline
    \multirow{3}{*}{PCPA \cite{kotseruba2021benchmark}} 
    & JAAD & \textcolor{blue}{0.58} & 0.50 & 0.71 & 0.61 & 0.58 \\
    & PIE & 0.87 & \textcolor{teal}{0.86} & 0.77 & 0.75 & \textcolor{teal}{0.79} \\
    & FU-PIP & 0.62 & 0.60 & \textcolor{purple}{0.58} & 0.51 & 0.47 \\
    \hdashline
    \multirow{2}{*}{CAPformer \cite{Lorenzo2021CAPformerPC}} 
    & PIE & 0.88 & 0.80 & 0.71 & 0.69 & 0.74 \\
    & FU-PIP & \textcolor{purple}{\underline{0.64}} & 0.60 & 0.55 & 0.58 & 0.54 \\
    \hdashline
    \multirow{3}{*}{GraphPlus \cite{cadena2022pedestrian}} 
    & JAAD & \textcolor{blue}{\textbf{0.70}} & \textcolor{blue}{\textbf{0.70}} & \textcolor{blue}{\textbf{0.76}} & \textcolor{blue}{\textbf{0.77}} & \textcolor{blue}{0.75} \\
    & PIE & \textcolor{teal}{0.89} & \textcolor{teal}{\textbf{0.90}} & \textcolor{teal}{0.81} & \textcolor{teal}{0.83} & \textcolor{teal}{0.79} \\
    & FU-PIP & \textcolor{purple}{0.64} & \textcolor{purple}{0.61} & 0.57 & \textcolor{purple}{0.59} & \textcolor{purple}{0.56} \\
    \hdashline
    \multirow{2}{*}{PIP-Net \cite{azarmi2024pip}} 
    & PIE & \textcolor{teal}{\textbf{0.91}} & \textcolor{teal}{\textbf{0.90}} & \textcolor{teal}{\textbf{0.84}} & \textcolor{teal}{\textbf{0.85}} & \textcolor{teal}{\textbf{0.84}} \\
    & FU-PIP & \textcolor{purple}{\textbf{0.73}} & \textcolor{purple}{\textbf{0.71}} & \textcolor{purple}{\textbf{0.69}} & \textcolor{purple}{\textbf{0.70}} & \textcolor{purple}{\textbf{0.68}} \\
    \noalign{\hrule height 1.2pt}
    
    \multicolumn{7}{c}{\textbf{Vision-Language Foundation Models}} \\
    
    \hline
    \multirow{1}{*}{GPT4V-PBP \cite{huang2024gpt}} 
    & JAAD & 0.57 & 0.61 & 0.65 & 0.82 & 0.54 \\
    \hdashline
    \multirow{1}{*}{GPT4V-PBP Skip \cite{huang2024gpt}} 
    & JAAD & 0.55 & 0.59 & 0.64 & 0.81 & 0.53 \\
    \hdashline
    \multirow{1}{*}{OmniPredict \cite{hamomnipredict}} 
    & JAAD & 0.67 & 0.65 & 0.65 & 0.66 & 0.65 \\
    \hline
    \multirow{3}{*}{LLaVA-Next (3B)} 
    & JAAD & 0.65 & 0.55 & 0.54 & 0.56 & 0.53 \\
    & PIE  & 0.72 & 0.62 & 0.61 & 0.63 & 0.60 \\
    & FU-PIP & 0.66 & 0.59 & 0.58 & 0.60 & 0.57 \\  
    \hdashline
    \multirow{3}{*}{LLaVA-Next (7B)} 
    & JAAD & 0.68 & 0.59 & 0.58 & 0.60 & 0.57 \\
    & PIE  & 0.75 & 0.66 & 0.65 & 0.67 & 0.64 \\
    & FU-PIP & 0.68 & 0.62 & 0.61 & 0.63 & 0.60 \\
    \hdashline
    \multirow{3}{*}{GPT-4 mini (8B)} 
    & JAAD & \textcolor{blue}{0.70} & \textcolor{blue}{0.62} & \textcolor{blue}{0.61} & \textcolor{blue}{0.63} & \textcolor{blue}{0.60} \\
    & PIE  & \textcolor{teal}{0.77} & \textcolor{teal}{0.69} & \textcolor{teal}{0.68} & \textcolor{teal}{0.70} & \textcolor{teal}{0.67} \\
    & FU-PIP & \textcolor{purple}{0.70} & \textcolor{purple}{0.64} &\textcolor{purple}{0.63} & \textcolor{purple}{0.65} & \textcolor{purple}{0.62} \\  
    \hdashline   
    \multirow{3}{*}{GPT-4V (1.8T)} 
    & JAAD & \textcolor{blue}{\textbf{0.74}} & \textcolor{blue}{\textbf{0.67}} & \textcolor{blue}{\textbf{0.66}} &\textcolor{blue}{\textbf{0.68}} & \textcolor{blue}{\textbf{0.65}} \\
    & PIE  & \textcolor{teal}{\textbf{0.81}} & \textcolor{teal}{\textbf{0.74}} & \textcolor{teal}{\textbf{0.73}} & \textcolor{teal}{\textbf{0.75}} & \textcolor{teal}{\textbf{0.72}} \\
    & FU-PIP & \textcolor{purple}{\textbf{0.74}} & \textcolor{purple}{\textbf{0.69}} & \textcolor{purple}{\textbf{0.68}} & \textcolor{purple}{\textbf{0.70}} & \textcolor{purple}{\textbf{0.67}} \\

    \noalign{\hrule height 1.2pt}
    \end{tabular}
    \label{tab:vlfm_comparison}
    \vspace{-1.5mm}
    {\scriptsize
    \begin{flushleft}
        Top values are indicated in \textbf{bold} and colour-coded as \textcolor{blue}{blue} for JAAD, \textcolor{teal}{teal} for PIE, and \textcolor{purple}{purple} for FU-PIP; second-top values are only colour-coded without bold style.
    \end{flushleft}}
\vspace{-0.7cm}
\end{table}

In contrast, the VLFMs, particularly the larger models, consistently outperform their vision-based counterparts. GPT-4V (1.8T) demonstrates superior performance across all datasets, achieving the highest accuracy (0.74 on JAAD, 0.81 on PIE, 0.74 on FU-PIP) and F1 scores (0.66 on JAAD, 0.73 on PIE, 0.68 on FU-PIP). This suggests that the incorporation of vision and language processing enables better contextual understanding and prediction, enhancing model effectiveness. GPT-4 mini (8B) and LLaVA-Next (7B) also show competitive performance, particularly on PIE, with GPT-4 mini achieving a high F1 score of 0.73. However, smaller models like LLaVA-Next (3 billion parameters) show lower performance, especially on the JAAD dataset, where it struggles with accuracy (0.65) and F1 (0.54), highlighting the limitations of smaller models in pedestrian intention prediction tasks.

In scenarios involving complex pedestrian behaviour, such as at designated environments like traffic lights and crosswalks, partial occlusion at intersections, or hesitation in parking lots, most VLFMs, particularly GPT-4V, demonstrate superior performance over conventional vision-based models. These models can reason about the relationships between traffic infrastructure, pedestrians, and surrounding vehicles, providing a more comprehensive and context-aware understanding of pedestrian situations and their potential crossing intentions.

\section{Conclusion}

This study investigated the adaptation of vision-language foundation models (VLFMs) for pedestrian crossing intention prediction, focusing on the impact of context-rich prompts and vehicle dynamics. Experimental results demonstrated that integrating hierarchical prompts and vehicle speed information significantly enhanced the prediction performance. Specifically, time-conscious representations of vehicle dynamics, which incorporate changes in speed over time, resulted in a 12.5\% increase in prediction accuracy compared to the baseline templates with no vehicle speed information. 
Body orientation- and pose-based prompts also 
contributed to an 8.7\% improvement, emphasising the importance of physical cues in predicting pedestrian intentions. 

The Automatic Prompt Engineer (APE) framework proved to be effective in optimising prompt design, achieving an additional 5.3\% gain in execution accuracy by refining linguistic and contextual diversity. To mitigate bias toward GPT-4V, we extended optimisation to other VLFMs for 20 iterations but found no further improvements in the top-10 prompts, suggesting performance differences stem from model architecture. Overall, the combination of contextually enriched prompts, vehicle dynamics, and systematic optimisation techniques enabled VLFMs to outperform previous works on VLFMs and most vision-based benchmark models in task-specific metrics.

\section*{Declarations}
The authors declare no conflicts of interest and no financial or personal relationships that could potentially influence the interpretation or presentation of the findings. 

\section*{Acknowledgment}
This research has received funding from the European Union’s Horizon 2020 research and innovation programme, for the Hi-Drive project under grant Agreement No 101006664. The article reflects only the author’s view, and neither the European Commission nor CINEA is responsible for any use that may be made of the information this document contains.

\bibliographystyle{IEEEtran}
\bibliography{ref.bib}

\begin{thebibliography}{10}
\providecommand{\url}[1]{#1}
\csname url@samestyle\endcsname
\providecommand{\newblock}{\relax}
\providecommand{\bibinfo}[2]{#2}
\providecommand{\BIBentrySTDinterwordspacing}{\spaceskip=0pt\relax}
\providecommand{\BIBentryALTinterwordstretchfactor}{4}
\providecommand{\BIBentryALTinterwordspacing}{\spaceskip=\fontdimen2\font plus
\BIBentryALTinterwordstretchfactor\fontdimen3\font minus \fontdimen4\font\relax}
\providecommand{\BIBforeignlanguage}[2]{{%
\expandafter\ifx\csname l@#1\endcsname\relax
\typeout{** WARNING: IEEEtran.bst: No hyphenation pattern has been}%
\typeout{** loaded for the language `#1'. Using the pattern for}%
\typeout{** the default language instead.}%
\else
\language=\csname l@#1\endcsname
\fi
#2}}
\providecommand{\BIBdecl}{\relax}
\BIBdecl

\bibitem{rasouli2019autonomous}
A.~Rasouli and J.~K. Tsotsos, ``Autonomous vehicles that interact with pedestrians: A survey of theory and practice,'' \emph{IEEE transactions on intelligent transportation systems}, vol.~21, no.~3, pp. 900--918, 2019.

\bibitem{razali2021pedestrian}
H.~Razali, T.~Mordan, and A.~Alahi, ``Pedestrian intention prediction: A convolutional bottom-up multi-task approach,'' \emph{Transportation research part C: emerging technologies}, vol. 130, p. 103259, 2021.

\bibitem{zhang2022st}
X.~Zhang, P.~Angeloudis, and Y.~Demiris, ``St crossingpose: A spatial-temporal graph convolutional network for skeleton-based pedestrian crossing intention prediction,'' \emph{IEEE Transactions on Intelligent Transportation Systems}, vol.~23, no.~11, pp. 20\,773--20\,782, 2022.

\bibitem{azarmi2024pip}
M.~Azarmi, M.~Rezaei, H.~Wang, and S.~Glaser, ``Pip-net: Pedestrian intention prediction in the wild,'' \emph{arXiv preprint arXiv:2402.12810}, 2024.

\bibitem{rasouli2020PedestrianAA}
A.~Rasouli, I.~Kotseruba, and J.~K. Tsotsos, ``Pedestrian action anticipation using contextual feature fusion in stacked rnns,'' in \emph{British Machine Vision Conference}, 2020.

\bibitem{kotseruba2020they}
I.~Kotseruba, A.~Rasouli, and J.~K. Tsotsos, ``Do they want to cross? understanding pedestrian intention for behavior prediction,'' in \emph{2020 IEEE Intelligent Vehicles Symposium (IV)}, 2020, pp. 1688--1693.

\bibitem{rasouli2022multi}
A.~Rasouli, T.~Yau, M.~Rohani, and J.~Luo, ``Multi-modal hybrid architecture for pedestrian action prediction,'' in \emph{2022 IEEE Intelligent Vehicles Symposium (IV)}.\hskip 1em plus 0.5em minus 0.4em\relax IEEE, 2022, pp. 91--97.

\bibitem{cadena2022pedestrian}
P.~R.~G. Cadena, Y.~Qian, C.~Wang, and M.~Yang, ``Pedestrian graph+: A fast pedestrian crossing prediction model based on graph convolutional networks,'' \emph{IEEE Transactions on Intelligent Transportation Systems}, vol.~23, no.~11, pp. 21\,050--21\,061, 2022.

\bibitem{chen2021visual}
T.~Chen, R.~Tian, and Z.~Ding, ``Visual reasoning using graph convolutional networks for predicting pedestrian crossing intention,'' in \emph{Proceedings of the IEEE/CVF international conference on computer vision}, 2021, pp. 3103--3109.

\bibitem{Lorenzo2021CAPformerPC}
J.~Lorenzo, I.~Parra, R.~Izquierdo, A.~L. Ballardini, {\'A}.~Hern{\'a}ndez-Saz, D.~F. Llorca, and M.~{\'A}. Sotelo, ``{CAPformer}: Pedestrian crossing action prediction using transformer,'' \emph{Sensors (Basel, Switzerland)}, vol.~21, 2021.

\bibitem{zhou2023pit}
Y.~Zhou, G.~Tan, R.~Zhong, Y.~Li, and C.~Gou, ``Pit: Progressive interaction transformer for pedestrian crossing intention prediction,'' \emph{IEEE Transactions on Intelligent Transportation Systems}, 2023.

\bibitem{sharma2025predicting}
N.~Sharma, C.~Dhiman, and S.~Indu, ``Predicting pedestrian intentions with multimodal intentformer: A co-learning approach,'' \emph{Pattern Recognition}, vol. 161, p. 111205, 2025.

\bibitem{yang2021pedestrian}
J.~Yang, A.~Gui, J.~Wang, and J.~Ma, ``Pedestrian behavior interpretation from pose estimation,'' in \emph{IEEE International Intelligent Transportation Systems Conference (ITSC)}, 2021, pp. 3110--3115.

\bibitem{ahmed2023multi}
S.~Ahmed, A.~Al~Bazi, C.~Saha, S.~Rajbhandari, and M.~N. Huda, ``Multi-scale pedestrian intent prediction using 3d joint information as spatio-temporal representation,'' \emph{Expert Systems With Applications}, vol. 225, p. 120077, 2023.

\bibitem{liu2020spatiotemporal}
B.~Liu, E.~Adeli, Z.~Cao, K.-H. Lee, A.~Shenoi, A.~Gaidon, and J.~C. Niebles, ``Spatiotemporal relationship reasoning for pedestrian intent prediction,'' \emph{IEEE Robotics and Automation Letters}, vol.~5, no.~2, pp. 3485--3492, 2020.

\bibitem{saleh2019real}
K.~Saleh, M.~Hossny, and S.~Nahavandi, ``Real-time intent prediction of pedestrians for autonomous ground vehicles via spatio-temporal densenet,'' in \emph{2019 International Conference on Robotics and Automation (ICRA)}, 2019, pp. 9704--9710.

\bibitem{xu2024pedestrian}
L.~Xu, S.~You, G.~He, and Y.~Li, ``Pedestrian-vehicle information modulation for pedestrian crossing intention prediction,'' \emph{IEEE Transactions on Intelligent Vehicles}, 2024.

\bibitem{zhang2024causal}
K.~Zhang, Q.~Sun, C.~Zhao, and Y.~Tang, ``Causal reasoning in typical computer vision tasks,'' \emph{Science China Technological Sciences}, vol.~67, no.~1, pp. 105--120, 2024.

\bibitem{zhou2024vision}
X.~Zhou, M.~Liu, E.~Yurtsever, B.~L. Zagar, W.~Zimmer, H.~Cao, and A.~C. Knoll, ``Vision language models in autonomous driving: A survey and outlook,'' \emph{IEEE Transactions on Intelligent Vehicles}, 2024.

\bibitem{huang2024gpt}
J.~Huang, P.~Jiang, A.~Gautam, and S.~Saripalli, ``Gpt-4v takes the wheel: Promises and challenges for pedestrian behavior prediction,'' in \emph{Proceedings of the AAAI Symposium Series}, vol.~3, no.~1, 2024, pp. 134--142.

\bibitem{hamomnipredict}
J.-S. Ham, J.~Huang, P.~Jiang, J.~Moon, Y.~Kwon, S.~Saripalli, and C.~Kim, ``Omnipredict: Gpt-4o enhanced multi-modal pedestrian crossing intention prediction.''

\bibitem{munirpedvlm}
\BIBentryALTinterwordspacing
F.~Munir, S.~Azam, T.~Mihaylova, V.~Kyrki, and T.~P. Kucner, ``Pedvlm: Pedestrian vision language model for intentions prediction,'' 2024. [Online]. Available: \url{https://openreview.net/forum?id=RAX45dcfA2}
\BIBentrySTDinterwordspacing

\bibitem{zhang2023cross}
C.~Zhang, A.~H. Kalantari, Y.~Yang, Z.~Ni, G.~Markkula, N.~Merat, and C.~Berger, ``Cross or wait? predicting pedestrian interaction outcomes at unsignalized crossings,'' in \emph{2023 IEEE Intelligent Vehicles Symposium (IV)}.\hskip 1em plus 0.5em minus 0.4em\relax IEEE, 2023, pp. 1--8.

\bibitem{azarmi2024feature}
M.~Azarmi, M.~Rezaei, H.~Wang, and A.~Arabian, ``Feature importance in pedestrian intention prediction: A context-aware review,'' \emph{arXiv preprint arXiv:2409.07645}, 2024.

\bibitem{budagam2024hierarchical}
D.~Budagam, S.~KJ, A.~Kumar, V.~Jain, and A.~Chadha, ``Hierarchical prompting taxonomy: A universal evaluation framework for large language models,'' \emph{arXiv preprint arXiv:2406.12644}, 2024.

\bibitem{zhou2022large}
Y.~Zhou, A.~I. Muresanu, Z.~Han, K.~Paster, S.~Pitis, H.~Chan, and J.~Ba, ``Large language models are human-level prompt engineers,'' \emph{arXiv preprint arXiv:2211.01910}, 2022.

\bibitem{kotseruba2021benchmark}
I.~Kotseruba and A.~Rasouli, ``Benchmark for evaluating pedestrian action prediction,'' \emph{IEEE Winter Conference on Applications of Computer Vision (WACV)}, pp. 1257--1267, 2021.

\bibitem{kong2023better}
A.~Kong, S.~Zhao, H.~Chen, Q.~Li, Y.~Qin, R.~Sun, X.~Zhou, E.~Wang, and X.~Dong, ``Better zero-shot reasoning with role-play prompting,'' \emph{arXiv preprint arXiv:2308.07702}, 2023.

\bibitem{zhang2024good}
C.~Zhang and S.~Wang, ``Good at captioning, bad at counting: Benchmarking gpt-4v on earth observation data,'' \emph{arXiv preprint arXiv:2401.17600}, 2024.

\bibitem{fang2018pedestrian}
Z.~Fang and A.~M. L{\'o}pez, ``Is the pedestrian going to cross? answering by 2d pose estimation,'' in \emph{2018 IEEE Intelligent Vehicles symposium (IV)}.\hskip 1em plus 0.5em minus 0.4em\relax IEEE, 2018, pp. 1271--1276.

\bibitem{OpenAI2023ChatGPT}
\BIBentryALTinterwordspacing
OpenAI, ``Chatgpt: Generative pre-trained transformer,'' 2023, accessed: 2025-01-18. [Online]. Available: \url{https://openai.com/chatgpt}
\BIBentrySTDinterwordspacing

\bibitem{Rasouli2017IV}
A.~Rasouli, I.~Kotseruba, and J.~K. Tsotsos, ``Agreeing to cross: How drivers and pedestrians communicate,'' in \emph{IEEE Intelligent Vehicles Symposium (IV)}, 2017, pp. 264--269.

\bibitem{rasouli2019pie}
A.~Rasouli, I.~Kotseruba, T.~Kunic, and J.~K. Tsotsos, ``{PIE}: A large-scale dataset and models for pedestrian intention estimation and trajectory prediction,'' in \emph{Proceedings of the IEEE/CVF International Conference on Computer Vision}, 2019, pp. 6262--6271.

\bibitem{achiam2024gpt}
J.~Achiam, S.~Adler, S.~Agarwal, L.~Ahmad, I.~Akkaya, F.~L. Aleman, D.~Almeida, J.~Altenschmidt, S.~Altman, S.~Anadkat \emph{et~al.}, ``Gpt-4 technical report,'' \emph{arXiv preprint arXiv:2303.08774}, 2024.

\bibitem{zhu2023minigpt}
D.~Zhu, J.~Chen, X.~Shen, X.~Li, and M.~Elhoseiny, ``Minigpt-4: Enhancing vision-language understanding with advanced large language models,'' \emph{arXiv preprint arXiv:2304.10592}, 2023.

\bibitem{li2024llava}
F.~Li, R.~Zhang, H.~Zhang, Y.~Zhang, B.~Li, W.~Li, Z.~Ma, and C.~Li, ``Llava-next-interleave: Tackling multi-image, video, and 3d in large multimodal models,'' \emph{arXiv preprint arXiv:2407.07895}, 2024.

\end{thebibliography}

\end{document}